\documentclass{article}

\PassOptionsToPackage{square,numbers,sort&compress}{natbib}
\usepackage{latex}

\usepackage[utf8]{inputenc} 
\usepackage[T1]{fontenc}    
\usepackage{hyperref}       
\usepackage{url}            
\usepackage{booktabs}       
\usepackage{amsfonts}       
\usepackage{nicefrac}       
\usepackage{microtype}      
\usepackage{xcolor}         
\usepackage{graphicx} 
\usepackage{subfigure}
\usepackage{bm}
\usepackage{amsmath}        
\usepackage{bbm}
\usepackage{amssymb}
\usepackage{algorithm}
\usepackage{algorithmic}
\usepackage{float}
\usepackage{multirow}
\usepackage{wrapfig}
\usepackage{graphicx}
\usepackage{subfigure}

\title{Learning Uniform Clusters on Hypersphere for Deep Graph-level Clustering}

\newcommand{\avector}{\frac{1}{N}\bm{1}}
\newcommand{\bvector}{\frac{1}{C}\bm{1}}
\newcommand{\modelname}{\textbf{UDGC}}

\author{
Mengling Hu, Chaochao Chen, Weiming Liu, XinyiZhang, Xinting Liao, and Xiaolin Zheng, \\
Zhejiang University, China\\
\texttt{\{humengling,zjuccc,21831010,22160177,xintingliao,xlzheng\}@zju.edu.cn}
}

\begin{document}

\maketitle

\begin{abstract}
Graph clustering has been popularly studied in recent years.
However, most existing graph clustering methods focus on node-level clustering, i.e., grouping nodes in a single graph into clusters.
In contrast, graph-level clustering, i.e., grouping multiple graphs into clusters, remains largely unexplored.
Graph-level clustering is critical in a variety of real-world applications, such as, properties prediction of molecules and community analysis in social networks.
However, graph-level clustering is challenging due to the insufficient discriminability of graph-level representations, and the insufficient discriminability makes deep clustering be more likely to obtain degenerate solutions (cluster collapse).
To address the issue, we propose a novel deep graph-level clustering method called Uniform Deep Graph Clustering (\modelname).
\modelname~assigns instances evenly to different clusters and then scatters those clusters on unit hypersphere, leading to a more uniform cluster-level distribution and a slighter cluster collapse.
Specifically, we first propose Augmentation-Consensus Optimal Transport (ACOT) for generating uniformly distributed and reliable pseudo labels for partitioning clusters.
Then we adopt contrastive learning to scatter those clusters.
Besides, we propose Center Alignment Optimal Transport (CAOT) for guiding the model to learn better parameters, which further promotes the cluster performance.
Our empirical study on eight well-known datasets demonstrates that \modelname~significantly outperforms the state-of-the-art models.
\end{abstract}
\section{Introduction}
Deep graph clustering becomes more and more attractive in recent years \cite{wang2019AGC,bo2020,tu2021deep,liu2022deep,yang2023cluster}, with the development of graph neural networks \cite{gcn2017,2019gin,wu2020comprehensive} and deep clustering \cite{xie2016unsupervised,jiang2017,caron2018deep,zhang2021supporting}.
%
Although good performance has been achieved, we found that most existing deep graph clustering models focus on node-level clustering, i.e., grouping nodes in \textit{a single graph} into clusters.
The work on graph-level clustering, i.e., grouping \textit{multiple graphs} into clusters, remains largely unexplored.
Graph-level clustering has a variety of real-world applications, for example, protein clustering can be used to construct meaning and stable groups of similar proteins for analysis and functional annotation \cite{zaslavsky2016clustering}.

However, deep graph-level clustering is challenging, the reasons are as follows.
First, the graph-level representations are usually obtained via performing a global pooling (average, sum, etc.) on all node-level representations, which definitely loses information over graphs.
Second, unlike node-level clustering that we can derive extra supervision for each node from their neighbors via propagation, graph-level clustering has no extra supervision signal.
As a result, the discriminability of graph-level representations is limited, and thus performing deep clustering on graphs will more likely to obtain the trivial solution that all representations converge to a single point (representation collapse) \cite{zhou2022comprehensive}.
%
A recently published deep graph-level clustering work GLCC \cite{ju2022glcc} tries to solve the problem by utilizing contrastive learning to learn discriminative and cluster-friendly graph-level representations with the supervision of pseudo labels.
GLCC can improve the discriminability of graph-level representations and avoid representation collapse to some extent.
However, GLCC utilizes the augmentation in view of diverse nature of graphs (e.g., node dropping, edge perturbation, and etc) which cannot preserve semantics well \cite{sun2021mocl,li2022let,xia2022simgrace}, and thus degrading the effectiveness of contrastive learning.
Besides, GLCC simply selects some predictions with smaller entropy values to generate pseudo labels, which may assign a specific label to many samples.
%
Thus, GLCC is easy to sink into degenerate solutions where many instances are assigned to a single cluster and most of clusters have only few samples or even no sample (cluster collapse).
In summary, the main challenge is \textit{How to deal with cluster collapse and enhance deep graph-level clustering performance.}

To address the aforementioned issues, in this paper, we propose Uniform Deep Graph Clustering (\modelname), a novel end-to-end framework for graph-level clustering.
To begin with, inspired by the observation that graph data can preserve their semantics well during encoder perturbations \cite{xia2022simgrace}, we propose to utilize a perturbed graph encoder for graph augmentation.
Then, we devise two modules in \modelname, i.e., \textit{pseudo label generation module} and \textit{representation enhancement module}.
The pseudo label generation module generates uniformly distributed pseudo labels for original graph representations and augmented graph representations.
The representation enhancement module uses the generated pseudo labels as supervision to facilitate intra-cluster compactness and inter-cluster separability.
Overall, we optimize \modelname~by expectation-maximization algorithm, i.e., we iteratively perform E-step (generating pseudo labels by the pseudo label generation module) and M-step (training the model by representation enhancement module).

The key idea of \modelname~to deal with cluster collapse is to provide cluster-level \textit{uniformity} of the features on hypersphere.
The cluster-level uniformity consists of two aspects:
1. instances are uniformly assigned to different clusters and
2. different clusters are uniformly distributed on hypersphere.
%
%
%
%
%
%
%
%
%
Specifically, we first propose Augmentation-Consensus Optimal Transport (ACOT) in the pseudo label generation module for generating pseudo labels to uniformly partition instances.
Due to graphs can be noisy, incomplete, or uncertain, ACOT also enhances the robustness of pseudo labels.
That is, we obtain two cost matrixes for original graph representations and augmented graph representations respectively, and assume that the corresponding two transport matrixes should be similar, which brings reliable pseudo labels that can be invariant to unneeded noise factors.
%
%
Note that, the experiment results show that our method with uniformly distributed pseudo labels can achieve better performance on both balanced and imbalanced graph datasets compared with existing graph-level clustering methods.
Then, we adopt contrastive learning in the representation enhancement module to scatter those clusters on unit hypersphere.
Besides, we propose Center Alignment Optimal Transport (CAOT) in the representation enhancement module to enhance the consensus of representation space and model parameter space, for guiding the model to learn better parameters and further enhancing clustering performance.

We summarize our main contributions as follows:
(1) We propose a novel end-to-end framework, i.e., \modelname, for graph-level clustering, the key idea is to provide cluster-level \textit{uniformity} of the features on hypersphere for coping with cluster collapse problem.
(2) To our best knowledge, this is the first attempt to propose Augmentation-Consensus Optimal Transport (ACOT) for generating pseudo labels, which provides robustness to unneeded noise factors.
(3) We propose Center Alignment Optimal Transport (CAOT) to enhance the consensus of representation space and model parameter space, which is benefit to learn better model parameters and obtain better clustering performance.
(4) We conduct extensive experiments on eight popularly used graph datasets and the results demonstrate the superiority of \modelname.

\section{Methodology}
\subsection{Overview}
\begin{figure}
  \centering
  \includegraphics[width=1\linewidth]{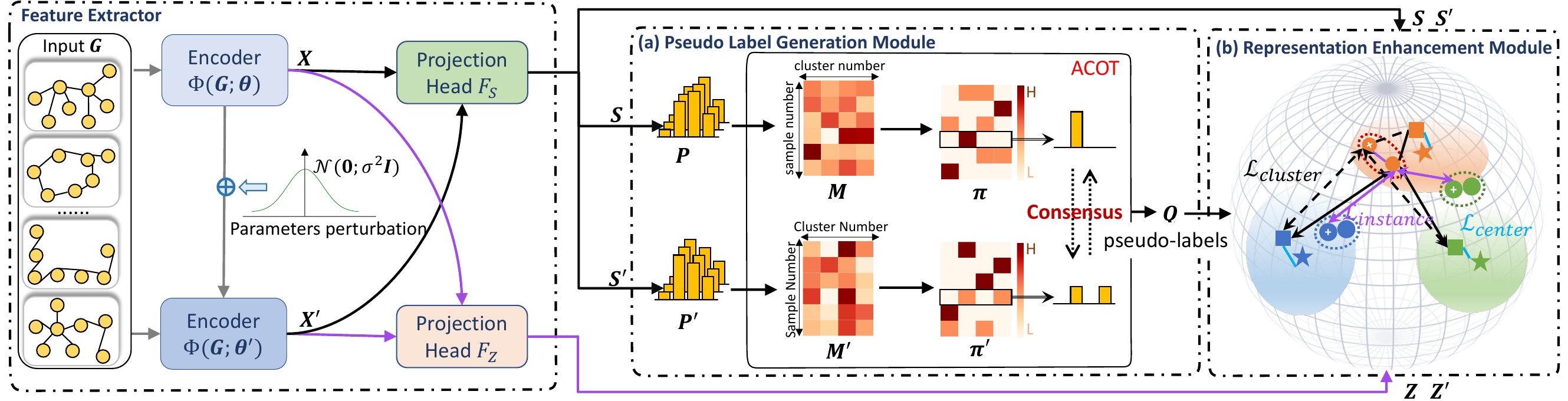}
  \vspace{-0.4cm}
  \caption{{The framework of \modelname.
  %
  %
  }
  }
  \vspace{-0.3cm}
  \label{fig:model}
\end{figure}

The goal of deep graph-level clustering is to learn the representations of graphs and perform clustering simultaneously. 
Our proposed Uniform Deep Graph Clustering (\modelname)~consists of \textit{pseudo labels generation module} and \textit{representation enhancement module}, as illustrated in Fig.\ref{fig:model}.
The pseudo label generation module aims to generate uniformly distributed and reliable pseudo labels.
To achieve this goal, we propose Augmentation-Consensus Optimal Transport (ACOT) to excavate pseudo label information from the predictions of original graph representations and augmented graph representations.
The representation enhancement module aims to use the pseudo labels as supervision to learn clustering-friendly representations while promoting cluster-level uniformity.
To achieve this goal, we introduce instance-based representation enhancement with contrastive learning and center-based representation enhancement with Center Alignment Optimal Transport (CAOT).
%
%
%
%
In this way, \modelname~can deal with cluster collapse problem and enhance deep graph-level clustering performance.

\subsection{Feature extractor}
The feature extractor captures hyperspherical graph embeddings for the subsequent pseudo label generation module and the representation enhancement module, as shown in the left part of Fig.\ref{fig:model}.
We will introduce the feature extractor in two parts, i.e., instance augmentation and hyperspherical features.

\paragraph{Instance augmentation.}
%
%
Augmentation in view of diverse nature of graphs (e.g., node dropping, edge perturbation, and etc) cannot preserve semantics well \cite{li2022let,sun2021mocl,xia2022simgrace}.
Inspired by the observation that graph data can preserve their semantics well during encoder perturbations \cite{xia2022simgrace}, we utilize a perturbed graph encoder for graph augmentation, as shown in the left part of Fig.\ref{fig:model}.
The parameter perturbation can be formulated as $\bm{\theta}^{'}=\bm{\theta}+\sigma\bm{\epsilon}$ where $\bm{\epsilon}\sim \mathcal{N}(0,\bm{I})$, $\bm{\bm{\theta}^{'}}$ and $\bm{\theta}$ are the model parameters, and $\sigma$ denotes a perturbation hyper-parameter.
We feed the original $N$ graphs $\bm{G}$ into an encoder $\Phi(\bm{G};\bm{\theta})$ and its perturbed version $\Phi(\bm{G};\bm{\theta^{'}})$ to obtain graph representations $\bm{X}\in\mathbbm{R}^{N\times D}$ and the augmented graph representations $\bm{X^{'}}\in\mathbbm{R}^{N\times D}$, where $N$ is the sample number and $D$ is the dimension of the representations.
%
\paragraph{Hyperspherical features.}
We utilize a one-layer fully connected layer as the projection head $F_S$ to map representations to a temporary space and then perform $l_2$ normalization to generate hyperspherical features, i.e., $\frac{F_S(\bm{X})}{||F_S(\bm{X})||_2}=\bm{S}\in\mathbbm{R}^{N\times D}$ and $\frac{F_S(\bm{X}^{'})}{||F_S(\bm{X}^{'})||_2}=\bm{S}^{'}\in\mathbbm{R}^{N\times D}$.
Hyperspherical features $\bm{S}$ and $\bm{S}^{'}$ are used for cluster-level training.
We next utilize a two-layer fully connected network as the projection head $F_Z$ to map the representations to a temporary space and then perform $l_2$ normalization to generate hyperspherical features, i.e., $\frac{F_Z(\bm{X})}{||F_Z(\bm{X})||_2}=\bm{Z}\in\mathbbm{R}^{N\times D}$ and $\frac{F_Z(\bm{X}^{'})}{||F_Z(\bm{X}^{'})||_2}=\bm{Z}^{'}\in\mathbbm{R}^{N\times D}$.
Hyperspherical features $\bm{Z}$ and $\bm{Z}^{'}$ are used for instance-level training.

\subsection{Pseudo label generation module}
We then introduce the pseudo label generation module.
Pseudo labels can provide supervision information for deep clustering and promote more discriminative representations \cite{hu2021learning}.
GLCC \cite{ju2022glcc} simply selects some predictions with smaller entropy values to generate pseudo labels, which may assign a specific label to many samples.
Thus, GLCC is easy to sink into degenerate solutions.
To provide reliable supervision information and deal with the cluster collapse problem, we propose pseudo label generation module, which consists of two parts: hyperspherical predictions and Augmentation-Consensus Optimal Transport (ACOT), as shown in Fig.\ref{fig:model}(a).
ACOT encourages pseudo labels generated by original instances or augmented instances to be consistent for the reliability, and encourages pseudo labels to uniformly partition different clusters for dealing with the collapse problem.

\paragraph{Hyperspherical predictions.}
The cluster predictions are obtained according to the angles between the hyperspherical features and the normalized randomly initialized agents \cite{wang2017normface}.
We denote the agents as $\bm{W}\in \mathbbm{R}^{C\times D}$, where $C$ is the cluster number.
The predictions on unit hypersphere can be calculated as $\bm{P}=softmax(\bm{S}\bm{W}^T/\tau)\in\mathbbm{R}^{N\times C}$ and $\bm{P}^{'}=softmax(\bm{S}^{'}\bm{W}^T/\tau)\in\mathbbm{R}^{N\times C}$, where $\tau$ is the temperature parameter \cite{wang2017normface}.
%

\paragraph{ACOT.}
ACOT aims to exploit the cluster predictions of both original graph representations and augmented graph representations to discover reliable and uniformly distributed pseudo labels.
Our inspiration of ACOT comes from self-labeling \cite{asano2020self} which extends standard cross-entropy minimization to an optimal transport (OT) problem \cite{peyre2019computational}.
In this way, we can generate uniformly distributed pseudo-labels for preventing cluster collapse.
%
However, self-labeling adopts only one view of each sample as input, the calculated pseudo labels cannot be robust to noise, so the reliability of the generated pseudo labels cannot be guaranteed, especially on graphs that are noisy, incomplete, or uncertain.
Thus, we propose ACOT, a new paradigm of optimal transport, which utilizes the information from both original features and augmented features (two views of each sample) to generate reliable pseudo labels that can be invariant to unneeded noise factors.

We will provide the details of ACOT below.
Generally, the cost matrix is defined as $-\log(\bm{P})$ \cite{asano2020self} such that predictions with higher confidence enjoy a lower cost.
However, the definition may cause the cost values to be $\infty$, and consequently making trouble for the optimization process.
We denote the cost matrixes as $\bm{M}=\exp(-\bm{P})$ and $\bm{M}^{'}=\exp(-\bm{P}^{'})$.
Utilizing $\exp(-\bm{P})$ to calculate cost matrix can limit the cost values to a reasonable range, benefiting optimization process.
We denote the transport matrix of original features as $\bm{\pi}\in\mathbbm{R}^{N\times C}$, the transport matrix of augmented features as $\bm{\pi}^{'}\in\mathbbm{R}^{N\times C}$.
To incorporate the information from both original features $\bm{X}$ and augmented features $\bm{X}^{'}$, we utilize two corresponding cost matrixes $\bm{M}$ and $\bm{M}^{'}$.
We encourage the consensus of the transport matrix $\bm{\pi}$ and the transport matrix $\bm{\pi}^{'}$.
The idea is that the pseudo labels generated by optimal transport for original features and augmented features should be invariant, for providing the robustness of pseudo labels to unneeded noise.
Thus, we formulate ACOT optimization problem as:
\begin{equation}
    \begin{aligned}
        &\min_{\bm{\pi},\bm{\pi}^{'}}{\langle \bm{\pi}, \bm{M} \rangle}+\epsilon {\rm{KL}(\bm{\pi}||\bm{\pi}^{'})}+\langle \bm{\pi}^{'},\bm{M}^{'}\rangle +\epsilon{\rm{KL}(\bm{\pi}^{'}||\bm{\pi})}\\
        &s.t.\,\,\bm{\pi}\bm{1}=\frac{1}{N}\bm{1},\bm{\pi}^T\bm{1}=\frac{1}{C}\bm{1},\bm{\pi}\geq 0,
        \bm{\pi}^{'}\bm{1}=\frac{1}{N}\bm{1},\bm{\pi}^{'T}\bm{1}=\frac{1}{C}\bm{1},\bm{\pi}^{'}\geq 0,
    \end{aligned}
    \label{eq:dACOT}
\end{equation}
where $\epsilon$ is a hyper-parameter.
We adopt bidirectional KL-divergence for providing the consensus of the two transport matrixes.
%
%
As we choose $\bm{\pi}$ to calculate the pseudo labels in the end, $\bm{\pi}^{'}$ can be regarded as a mediator to prompt $\bm{\pi}$ for utilizing the commonplace of $\bm{M}$ and $\bm{M}^{'}$, i.e., the unneeded noise can be ignored.
Furthermore, ACOT promotes the cluster assignment solution $\bm{\pi}$ to evenly assign $N$ samples to $C$ clusters, which deals with cluster collapse problem.
The optimization objective can be tractably solved by the Sinkhorn-Knopp \cite{cuturi2013sinkhorn} style algorithm, which needs only a few computational overheads.

\begin{wrapfigure}{r}{0.50\textwidth}
\vspace{-24pt}
\begin{center}
\includegraphics[width=0.45\textwidth]{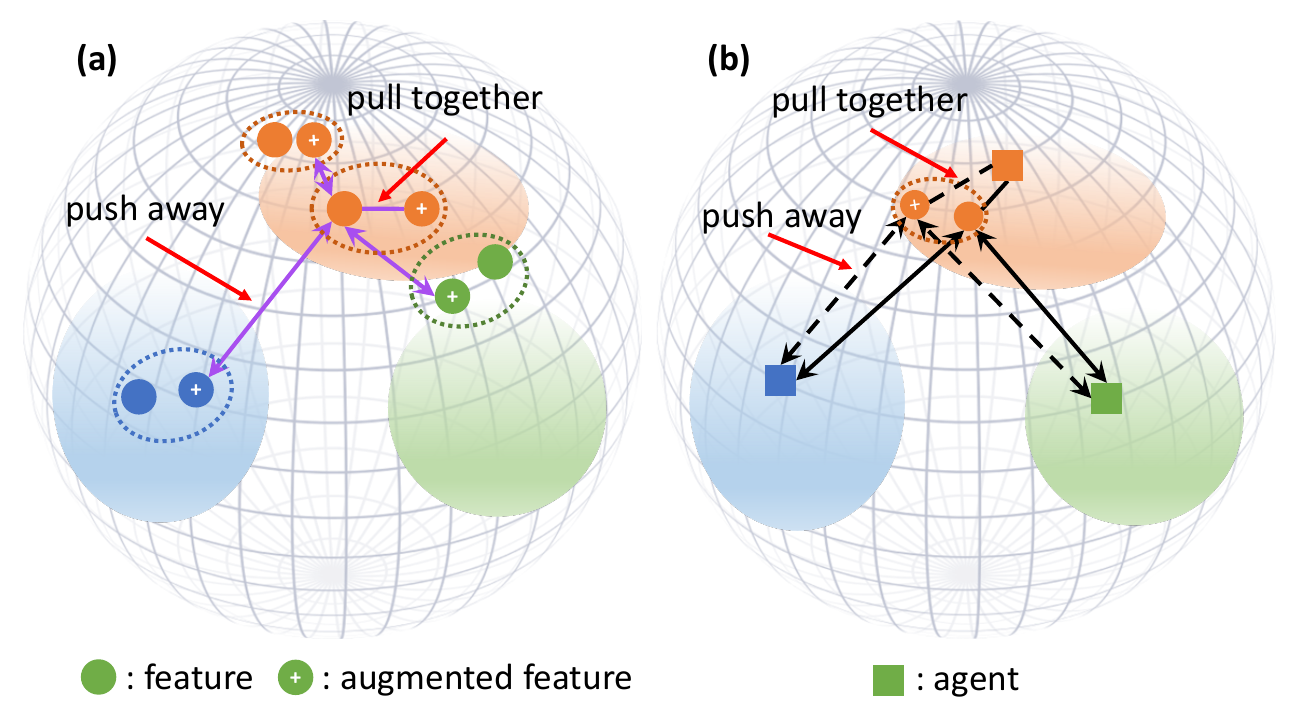}
\end{center}
\vspace{-8pt}
\caption{Illustration of instance-based representation enhancement: (a) instance-instance representation enhancement, (b) instance-agent representation enhancement.}
\label{fig:instance}
\vspace{-28pt}
\end{wrapfigure}

%
We provide the optimization details in Appendix A.
After obtaining $\bm{\pi}$, we can get pseudo labels $\bm{q}\in\mathbbm{R}^{N}$ by argmax operation, i.e., $q_i={\mathrm{argmax}}_j\bm{\pi}_{ij}$.
Through the steps of hyperspherical predictions and augmentation-consensus optimal transport, we can generate reliable and uniformly distributed pseudo labels for the representation enhancement module.

\subsection{Representation enhancement module}

We now introduce the representation enhancement module.
To learn clustering-friendly representations which have small intra-cluster distance and large inter-cluster distance, we propose representation enhancement module, which consists of instance-based representation enhancement and center-based representation enhancement.
%
Instance-based representation enhancement aims to utilize the pseudo labels as supervision to encourage different clusters to be uniformly distributed on unit hypersphere and instances in the same cluster to be gathered.
Center-based representation enhancement aims to provide the consistency of representation space and model parameter space, which further promotes the clustering performance.

\subsubsection{Instance-based representation enhancement}

Instance-based representation enhancement consists of instance-instance representation enhancement and instance-agent representation enhancement.
Instance-instance representation enhancement is supposed to separate the overlapped clusters with instance-level uniformity on unit hypersphere.
While instance-agent representation enhancement is supposed to facilitate intra-cluster compactness and inter-cluster separability with cluster-level uniformity on unit hypersphere.

\paragraph{Instance-instance representation enhancement.}
We utilize contrastive learning for instance-instance representation enhancement, which pulls a feature and its augmented feature together while pushes away the feature from all the other augmented features, as shown in Fig.\ref{fig:instance}(a).
The instance-instance enhancement loss can be defined as:
\begin{equation}
    \begin{aligned}
    \mathcal{L}_{instance}&=\frac{1}{N}\sum_{n=1}^{N}-\log\frac{\exp(\frac{\bm{z}_n^T\bm{z}^{'}_n}{\tau})}{\sum_{i=1,i\neq n}^{N}\exp(\frac{\bm{z}_n^T\bm{z}_i^{'}}{\tau})}\\
    &=\underbrace{\frac{1}{N}\sum_{n=1}^N-\frac{\bm{z}_n^T\bm{z}_n^{'}}{\tau}}_{instance\,alignment} + \underbrace{\frac{1}{N}\sum_{n=1}^{N}\log{\sum_{i=1,i\neq n}^{N}\exp(\frac{\bm{z}_n^T\bm{z}_n^{'}}{\tau})}}_{instance\,uniformity},
    \end{aligned}
\label{eq:instance}
\end{equation}
where $\bm{z}_n$ is the $n_{th}$ row of features $\bm{Z}$, and $\bm{z}_n^{'}$ is the $n_{th}$ row of features $\bm{Z}^{'}$.
The loss in Eq.\ref{eq:instance} consists of \textit{instance alignment} and \textit{instance uniformity}.
The instance alignment encourages that the learned features can be invariant to unneeded noise factors.
The instance uniformity encourages the features uniformly distributed on unit hypersphere, and separates the overlapped clusters.

\paragraph{Instance-agent representation enhancement}
%
%
This part we first partition instances into different clusters with the generated pseudo labels and obtain the corresponding agent of each instance.
Then, we utilize contrastive learning to pulls each instance and its augmentation to the corresponding agent and pushes them away from all the other agents, as shown in Fig.\ref{fig:instance}(b).
The instance-agent enhancement loss for original features $\bm{S}$ is defined as below:
\begin{equation}
    \begin{aligned}
    \mathcal{L}_{agent}(\bm{S})&=\frac{1}{N}\sum_{n=1}^{N}-\log\frac{\exp(\frac{\bm{s}_n^T\bm{w}_{q_n}}{\tau})}{\sum_{i=1,i\neq q_n}^{C}\exp(\frac{\bm{s}_n^T\bm{w}_i}{\tau})}\\
    &=\underbrace{\frac{1}{N}\sum_{n=1}^N-\frac{\bm{s}_n^T\bm{w}_{q_n}}{\tau}}_{cluster\,alignment} + \underbrace{\frac{1}{N}\sum_{n=1}^{N}\log{\sum_{i=1,i\neq q_n}^{C}\exp(\frac{\bm{s}_n^T\bm{w}_i}{\tau})}}_{cluster\,uniformity},
    \end{aligned}
\label{eq:cluster}
\end{equation}
where $\bm{s}_n$ is the $n_{th}$ row of features $\bm{S}$, $\bm{w}_i$ is the $i_{th}$ row of agents $\bm{W}$, $q_n$ is the $n_{th}$ value in pseudo labels $\bm{q}$, and $\bm{w}_{q_n}$ denotes the corresponding agent of $n_{th}$ instance.
The loss in Eq.\ref{eq:cluster} consists of \textit{cluster alignment} and \textit{cluster uniformity}.
The cluster alignment encourages that instances with the same pseudo label to be clustered.
The cluster uniformity encourages that different clusters to be scattered on unit hypersphere, benefiting inter-cluster separability.
%
%
With the instance-agent enhancement loss for augmented features $\bm{S}^{'}$, the overall cluster instance-agent enhancement loss is defined as $\mathcal{L}_{agent}=\mathcal{L}_{agent}(\bm{S})+\mathcal{L}_{agent}(\bm{S}^{'})$.

\subsubsection{Center-based representation enhancement}
\begin{wrapfigure}{r}{0.50\textwidth}
\vspace{-24pt}
\begin{center}
\includegraphics[width=0.45\textwidth]{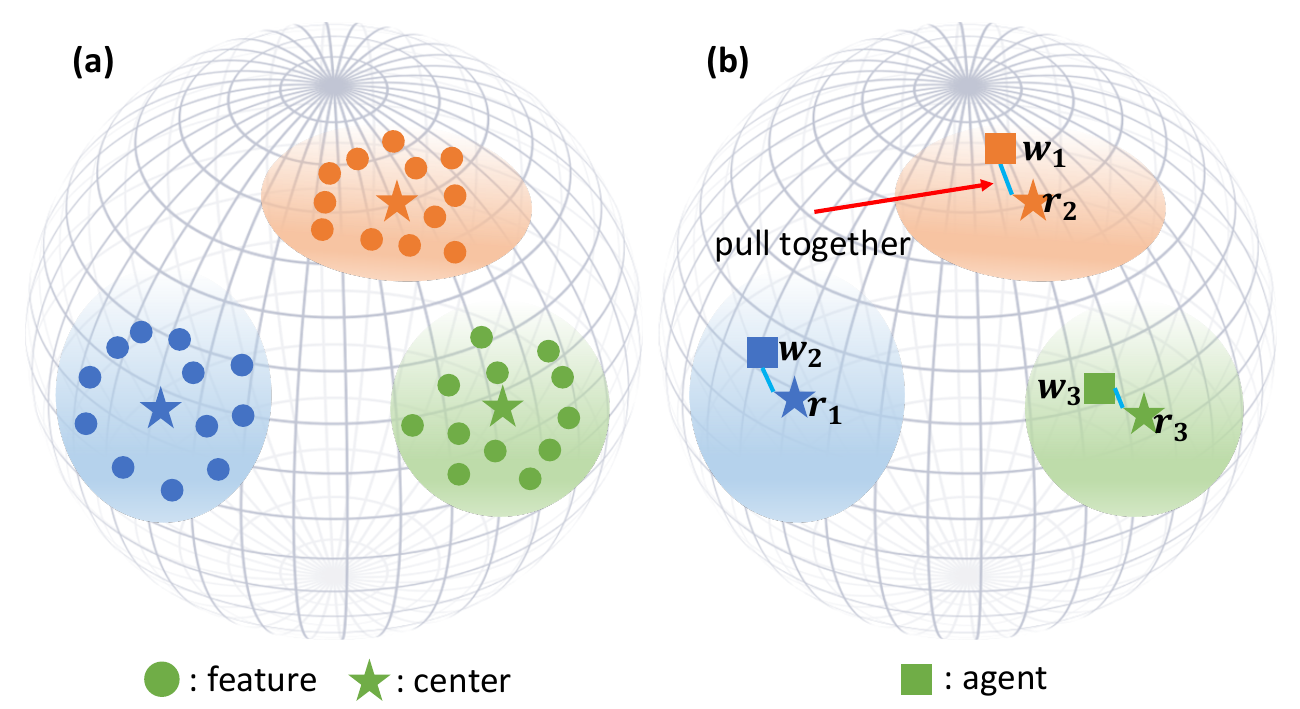}
\end{center}
\vspace{-8pt}
\caption{Illustration of center-base representation enhancement: (a) center discovery, (b) center alignment.}
\label{fig:center}
\vspace{-28pt}
\end{wrapfigure}

Center-based representation enhancement aims to enhance the consensus of representation space and model parameter space.
To achieve this goal, we propose Center Alignment Optimal Transport (CAOT).
CAOT consists of two steps: \textit{centers discovery} and \textit{center alignment}.
Specifically, CAOT first discovers centers in the representation space and then aligns the agents with the corresponding centers, as shown in Fig.\ref{fig:center}(a)-(b).
Note that agents are in model parameter space and can be regarded as trainable centers.
Aligning the centers from the representation space and the model parameter space can enhance the consensus of the two space.
The model parameter space cannot fully exploit the semantic information of samples, which limits the model fitting ability, leading to under-fitting.
While the representation space can utilize metric learning to fully exploit semantic information.
Thus, enhancing the consensus makes the representation space guide the model parameter space to learn better parameters, which enhances clustering performance.

\paragraph{Centers discovery.}
This step aims to discover centers in the representation space.
Specifically, we denote the centers of $\bm{X}$ as $\bm{\mu}\in \mathbbm{R}^{C\times D}$.
The centers should be different from each other and belong to different clusters.
To avoid the solution that all instances be clustered into few centers, we adopt optimal transport to make these instances uniformly gathered by all centers as:
\begin{equation}
    \begin{aligned}
        &\min_{\bm{\xi},\bm{\mu}}{\sum_{i=1}^{N}\sum_{j=1}^{C}\xi_{ij}||\bm{x}_i-\bm{\mu}_j||_2^2+\eta\bm{\xi}\log\bm{\xi}}
        \,\,s.t.\,\,\bm{\xi}\bm{1}=\frac{1}{N}\bm{1},\bm{\xi}^T\bm{1}=\frac{1}{C}\bm{1},\bm{\xi}\geq 0,
    \end{aligned}
    \label{eq:mu}
\end{equation}
where $\bm{\xi}$ is the transport matrix, $\eta$ is a hyper parameter to balance the coupling matching and the entropy regularization.
The problem can be optimized via alternatively update $\bm{\xi}$ and $\bm{\mu}$. We first fix $\bm{\mu}$ and update $\bm{\xi}$ via Sinkhorn algorithm \cite{cuturi2013sinkhorn}.
Then we fix $\bm{\xi}$ and update $\bm{\mu}_j$ to obtain the results as $\bm{\mu}_j=\sum_{i=1}^N\xi_{ij}\bm{x}_i/\sum_{i=1}^N\xi_{ij}$.
We can iteratively update them to obtain the convergent result.
In the end, we normalize $\bm{\mu}$ and obtain hyperspherical centers, i.e., $\bm{r}=\frac{\bm{\mu}}{||\bm{\mu}||_2}$.

\paragraph{Center alignment.}
After we obtain the hyperspherical centers in the representation space, we tend to align agents and the centers for enhancing the consensus.
Because clusters have no order, i.e., it is unknown each agent should be aligned to which center, we propose to match them by optimal transport first.
We formulate the matching problem as below,
\begin{equation}
    \begin{aligned}
        &\min_{\bm{\psi}}{\sum_{i=1}^{C}\sum_{j=1}^{C}\psi_{ij}(\exp(-\bm{w}_i\bm{r}_j^T))+\eta_1\bm{\psi}\log\bm{\psi}}
        \,\,s.t.\,\,\bm{\psi}\bm{1}=\frac{1}{N}\bm{1},\bm{\psi}^T\bm{1}=\frac{1}{C}\bm{1},\bm{\psi}\geq 0,
    \end{aligned}
    \label{eq:psi}
\end{equation}
where $\bm{\psi}$ is the transport matrix (i.e., matching solution), $\exp(-\bm{w}_i\bm{r}_j^T)$ is the cost to move mass from agent $\bm{w}_i$ to center $\bm{r}_j$ on unit hypersphere, $\eta_1$ is a hyper parameter to balance the coupling matching and the entropy regularization.
The agents and the centers don't correspond in order, take Fig.\ref{fig:center}(b) as an example, $\bm{w}_1$ matches to $\bm{r}_2$ and $\bm{w}_2$ matches to $\bm{r}_1$.
We use $\exp(-\bm{w}_i\bm{r}_j^T)$ as cost such that higher similarity between $\bm{w}_i$ and $\bm{r}_j$ enjoys a lower cost.
%
We can solve $\bm{\psi}$ via Sinkhorn algorithm \cite{cuturi2013sinkhorn}.
Then the center-agent representation enhancement loss can be defined as:
\begin{equation}
    \begin{aligned}
        \mathcal{L}_{center}={\sum_{i=1}^{C}\sum_{j=1}^{C}\psi_{ij}^*(\exp(-\bm{w}_i\bm{r}_j^T))},
    \end{aligned}
\end{equation}
where $\phi_{ij}^*$ is the optimal matching solution among agents and centers, higher ${\phi}_{ij}^*$ means that $\bm{w}_i$ and $\bm{r}_j$ are more likely to be matched.

In summary, we first generate the centers of representations $\bm{X}$ and use $l_2$ normalization to get the hyperspherical centers.
Then, we match the agents to their respective centers.
In the end, we propose the center-agent representation enhancement loss to align the agents to their corresponding centers.
The detailed optimization process of CAOT can be found in Appendix B.

\subsection{The optimization of \modelname}
The overall optimization of \modelname~is done by expectation-maximization algorithm.
Specifically, \textbf{E-step} aims to estimate pseudo labels by the pseudo label generation module.
We perform E-step throughout the whole training process in a logarithmic distribution \cite{asano2020self}.
\textbf{M-step} aims to minimize the objective of \modelname, which consists of instance-instance representation enhancement loss, instance-agent representation enhancement loss, and center-agent representation enhancement loss proposed in the representation enhancement module.
That is, the objective is given as:
\begin{equation}
    \begin{aligned}
\mathcal{L}=\lambda\mathcal{L}_{instance}+\mathcal{L}_{agent}+\mathcal{L}_{center},
    \end{aligned}
    \label{eq:all_loss}
\end{equation}
where $\lambda$ denotes the balanced hyper parameter.
By doing this, \modelname~can deal with cluster collapse problem and enhance deep graph-level clustering performance.
We provide the whole training process in detail in Appendix C.

\section{Empirical study}
\begin{table}
\caption{Experimental results on eight datasets}
\label{tab:results}
\centering

\begin{tabular}{lcccccccc}
\toprule
& \multicolumn{2}{c}{MUTAG}
& \multicolumn{2}{c}{DD}
& \multicolumn{2}{c}{COIL-DEL}
& \multicolumn{2}{c}{IMDB-B}\\
\cmidrule{2-9}
&ACC &NMI
&ACC &NMI
&ACC &NMI
&ACC &NMI\\

\midrule
Graphlet & 0.665 & 0.001 & 0.579 & 0.004  & 0.108 & 0.332 & 0.583 & 0.020\\ 
SP       & \textbf{0.830} & 0.299 & 0.585 & 0.003  & 0.117 & 0.373 & 0.567 & 0.035  \\ 
WL       & 0.702 & 0.078 & 0.586 & 0.006  & 0.113 & 0.357 & 0.530 & 0.023\\
\midrule
InfoGraph & 0.737  & 0.263 & 0.558 & 0.008  & 0.116 & 0.367  & 0.538 & 0.041 \\
GraphCL   & {0.785}  & \textbf{0.382} & 0.573 & 0.019 & 0.109 & 0.358 & 0.545 & 0.046 \\
CuCo      & 0.761  & 0.282 & 0.562 & 0.012 & 0.117 & 0.369  & 0.507 & 0.001\\
JOAO      & 0.771 & 0.316 & 0.578 & 0.012 & 0.113 & 0.367  & 0.543  & 0.042 \\
RGCL      & 0.677 & 0.076 & 0.565 & 0.014 & \underline{0.122} & 0.374  & 0.546 &  0.047 \\
SimGRACE  & 0.730 & 0.236 & 0.589 & 0.001 & 0.117 & \underline{0.377} & 0.559 &  0.049\\
\midrule 
GLCC      & 0.751 & 0.260 & \underline{0.607} & \underline{0.024} & 0.106 & 0.363 & \underline{0.665} & \underline{0.081}\\
\midrule
\modelname-V1  & 0.745 & 0.161 & 0.620 & 0.032 & 0.046 & 0.246 & 0.648 & 0.066\\
\modelname-V2 & 0.771 & 0.184 & 0.624 & 0.035 & 0.136 & 0.382 & 0.674 & 0.090\\
\modelname   & \underline{0.798} & \underline{0.353} & \textbf{0.637} & \textbf{0.041} & \textbf{0.141} & \textbf{0.387} & \textbf{0.687} & \textbf{0.104}\\
\midrule
\textbf{Improvement($\uparrow$)} & - & - & 4.942\% & 70.83\% & 15.57\% & 2.653\%&3.308\% & 28.40\%\\
\midrule
\end{tabular}

\begin{tabular}{lcccccccc}
& \multicolumn{2}{c}{REDDIT-B}
& \multicolumn{2}{c}{REDDIT-12K}
& \multicolumn{2}{c}{COLORS-3}
& \multicolumn{2}{c}{Syntie}\\

\cmidrule{2-9}
&ACC &NMI
&ACC &NMI
&ACC &NMI
&ACC &NMI\\

\midrule
Graphlet & 0.502 & 0.001 & 0.187 & 0.073 & >1Day & >1Day  & 0.353 & 0.080 \\ 
SP       & 0.577 & 0.021 & 0.204 & 0.062 & >1Day & >1Day  & 0.430 & 0.226 \\ 
WL       & 0.576 & 0.089 & 0.189 & 0.092 & >1Day & >1Day  & \textbf{0.545} & \underline{0.509} \\
\midrule
InfoGraph & 0.508 & 0.016 & 0.205 & 0.045   & \underline{0.111} & \underline{0.005} & {0.520} & {0.498} \\
GraphCL   & 0.519 & 0.033 & 0.181 & 0.096   & 0.110 & 0.004 & 0.492               & 0.449  \\
CuCo      & 0.510 & 0.018 & 0.192 & 0.003   & 0.109 & \underline{0.005} & 0.516      & 0.490  \\
JOAO      & 0.520 & 0.034 & 0.183 & 0.003  & 0.109 & 0.004 & 0.511      & 0.448\\
RGCL      & 0.509 & 0.017 & 0.092 & 0.003  & \underline{0.111} & \underline{0.005} & 0.519      & 0.493\\
SimGRACE  & 0.513 & 0.024 & 0.210 & 0.062  & 0.110 & 0.004 & 0.459      & 0.312 \\
\midrule 
GLCC      & \underline{0.676} & \underline{0.092} & \underline{0.226} & \underline{0.105}  & 0.109  & 0.004 & 0.507      & 0.429 \\
\midrule
\modelname-V1 & 0.655 & 0.113 & 0.236 & 0.082 & 0.105 & 0.003 & 0.451 & 0.291 \\
\modelname-V2 & 0.699 & 0.118 & 0.250 & 0.123 & 0.112 & 0.007 & 0.471 & 0.343 \\
\modelname  & \textbf{0.708} & \textbf{0.131} & \textbf{0.257} & \textbf{0.138} & \textbf{0.112} & \textbf{0.006} & \underline{0.525} & \textbf{0.569}\\
\midrule
\textbf{Improvement($\uparrow$)} & 4.734\% & 42.39\% & 13.72\% & 31.43\% &0.901\% & 20.00\% & - & 11.79\% \\
\bottomrule
\vspace{-0.4cm} 
\end{tabular}
\end{table}

In this section, we conduct experiments on eight well-known datasets to answer the following questions:
(1) \textbf{RQ1:} How does our approach perform compared with the state-of-the-art graph-level clustering methods?
(2) \textbf{RQ2:} How do ACOT and CAOT contribute to the performance improvement?
(3) \textbf{RQ3:} How does the performance of \modelname~vary with different values of the hyper-parameters?

\subsection{Experiment settings}
\paragraph{Datasets.}
We conduct extensive experiments on five kinds of datasets from TUDatasets \cite{tu2021deep}: Small molecules (MUTAG), Bioinformatics (DD), Computer vision (COIL-DEL), Social networks (IMDB-BINARY, REDDIT-BINARY, and REDDIT-MULTI-12K), and Synthetic (COLORS-3 and Synthie).
Among them, COIL-DEL, IMDB-BINARY, and REDDIT-BINARY are balanced datasets, other five datasets are imbalanced datasets.
More details about the datasets are shown in Appendix D.1.

\paragraph{Baselines.}
We compare our \modelname~with three graph kernel methods, six deep graph representation learning methods, and one deep graph-level clustering method.
Graph kernel methods include Graphlet \cite{shervashidze2009efficient}, Shortest Path Kernel (SP) \cite{borgwardt2005shortest}, and Weisfeiler-Lehman Kernel (WL) \cite{shervashidze2011weisfeiler}.
Deep graph representation learning methods include InfoGraph \cite{suninfograph}, GraphCL \cite{you2020graph}, CuCo \cite{chu2021cuco}, JOAO \cite{you2021graph}, RGCL \cite{li2022let}, and SimGRACE \cite{xia2022simgrace}.
For above methods, we first learn representations and then perform K-means \cite{macqueen1965some} to obtain the clustering results.
Note that there is only one deep graph-level clustering method for simultaneously representation learning and clustering, i.e., GLCC \cite{ju2022glcc}.

\paragraph{Implemented details.}
We build our model with PyTorch \cite{NEURIPS2019_9015} and train it using the Adam optimizer \cite{kingma2015adam}.
We set hyper-parameters $\eta$ and $\eta_1$ to $0.1$.
We study the effect of hyper-parameter $\epsilon$ on ACOT by varying it in $\{0, 0.05, 0.1, 0.2, 0.5\}$.
We study the effect of hyper-parameter $\lambda$ by varying it in $\{0, 0.01, 0.1, 1, 10\}$.
More details are provided in Appendix D.2.
Following previous graph clustering work \cite{bo2020,tu2021deep,liu2022deep,ju2022glcc}, we set the cluster numbers to the ground-truth category numbers.
We adopt three widely used metrics to evaluate the clustering performance, including accuracy (ACC) \cite{papadimitriou1998combinatorial}, normalized mutual information (NMI) \cite{vinh2009information} and adjusted rand index (ARI) \cite{steinley2004properties}.
%
%
For all the experiments, we repeat five times and report the average results.

\subsection{Clustering performance (RQ1)}
\paragraph{Results and discussion.}
The average results of ACC and NMI on eight datasets are shown in Table \ref{tab:results}.
%
More results can be found in Appendix D.3.
From them, we have the following observations:
(1) The graph kernel methods can only obtain good results on small datasets.
%
%
%
The reason is that larger datasets obtain higher dimensional representations which may contain too much irrelevant information, and it is more difficult for k-means to compare the similarity between higher dimensional representations, thus decreasing the clustering performance.
%
%
(2) Deep graph representation learning methods perform better than graph kernel methods on larger datasets (e.g., REDDIT-12K, COLORS-3, and COIL-DEL).
These methods solve the high dimensional representations problem of kernel methods by utilizing graph neural network to exploit low dimensional representations.
However, the overall clustering performance is still not good because the deep graph representation learning methods cannot learn clustering-friendly representations.
(3) GLCC obtains better results by utilizing contrastive learning with the supervision of pseudo labels to learn clustering-friendly representations.
However, GLCC utilizes the augmentation that cannot preserve semantics well, and thus degrading the effectiveness of contrastive learning.
Besides, GLCC simply selects some predictions with smaller entropy values to generate pseudo labels, which may assign a specific label to many samples.
Thus, GLCC tends to sink into degenerate solutions.
Overall, the clustering performance of GLCC is limited.
(4) Our method \modelname~outperforms all baselines, which proves that cluster-level uniformity can significantly improve the deep graph-level clustering performance.
%
%
%

\subsection{In-depth analysis (RQ2 and RQ3)}
\paragraph{Ablation (RQ2).}
To study how ACOT and CAOT contribute to the final performance, we compare \modelname~with its two variants, i.e., \modelname-V1 and \modelname-V2.
\modelname-V1 does not adopt ACOT and the pseudo labels are generated by traditional OT utilizing only one view of an instance.
\modelname-V2 dose not adopt CAOT, i.e., there is no center alignment loss.
The comparison results are shown in Table \ref{tab:results}.
From it, we can observe that:
(1) \modelname-V1 without ACOT obtain too noisy pseudo labels, resulting in poor performance.
(2) \modelname-V2 without CAOT achieves better performance than \modelname-V1 because \modelname-V2 utilizes ACOT to generate reliable pseudo labels. But it is worse than \modelname~since CAOT is missing.
(3) With ACOT and CAOT, \modelname~can obtain the best performance.
Overall, the above ablation study demonstrates that our proposed ACOT is effective in generating reliable pseudo labels, CAOT is effective in benefiting model training, both ACOT and CAOT are effective in deep graph-level clustering.

\begin{figure*} 
    \centering
    
    \subfigure[Effects of $\epsilon$]{
    \begin{minipage}[t]{0.23\linewidth} 
    \includegraphics[width=3.7cm]{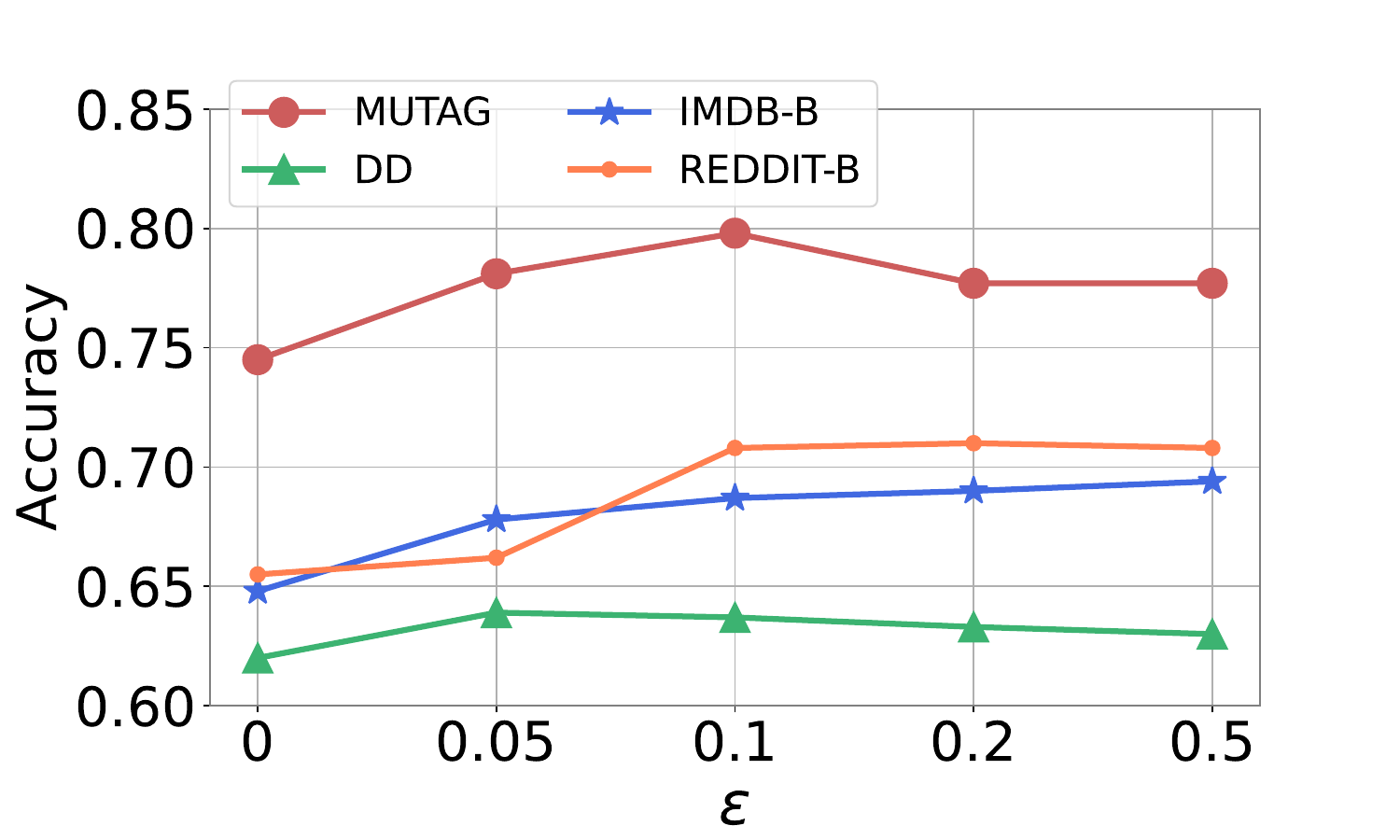}
    \end{minipage}
}
    \subfigure[Effects of $\epsilon$]{
    \begin{minipage}[t]{0.23\linewidth} 
    \includegraphics[width=3.7cm]{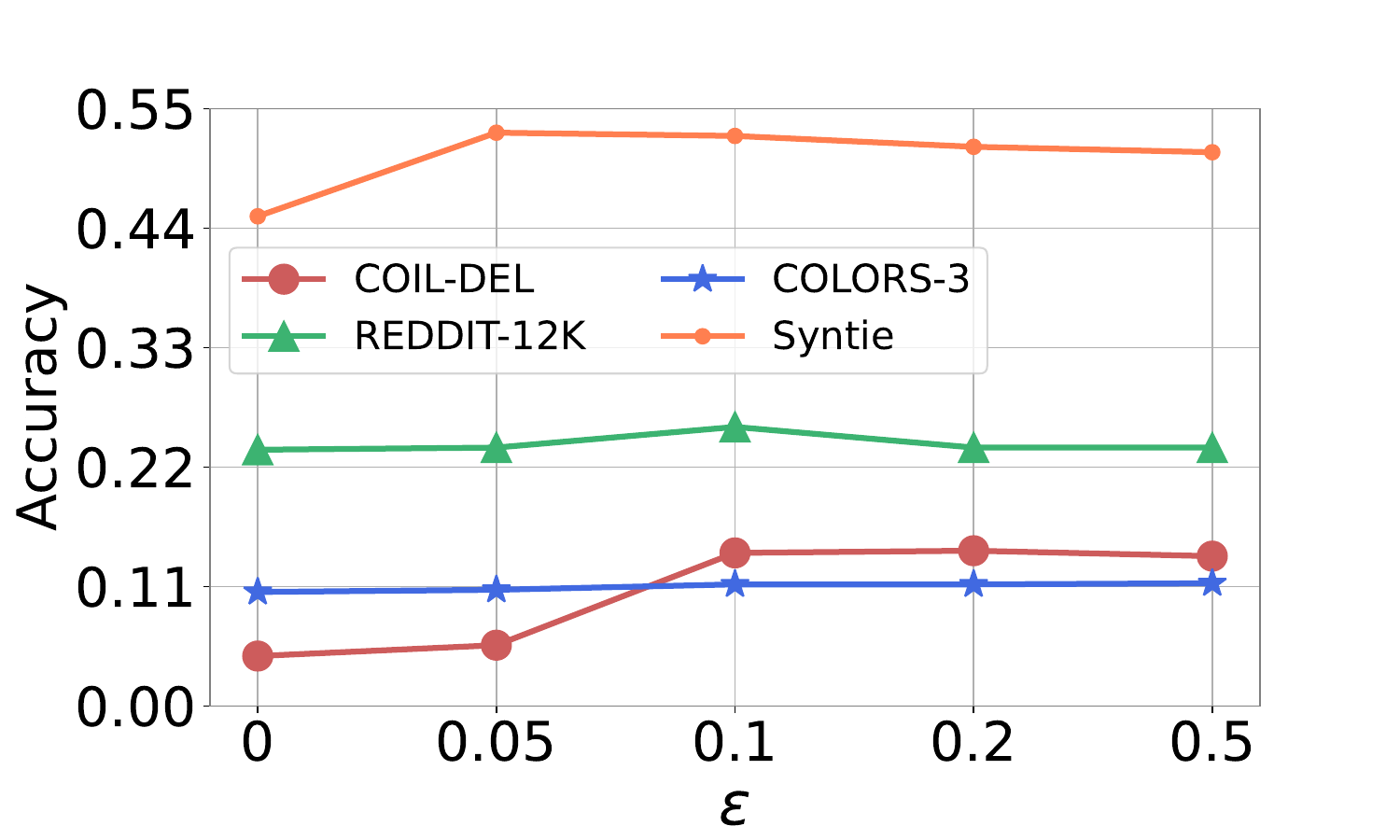}
    \end{minipage}
}
    \subfigure[Effects of $\lambda$]{
    \begin{minipage}[t]{0.23\linewidth} 
    \includegraphics[width=3.7cm]{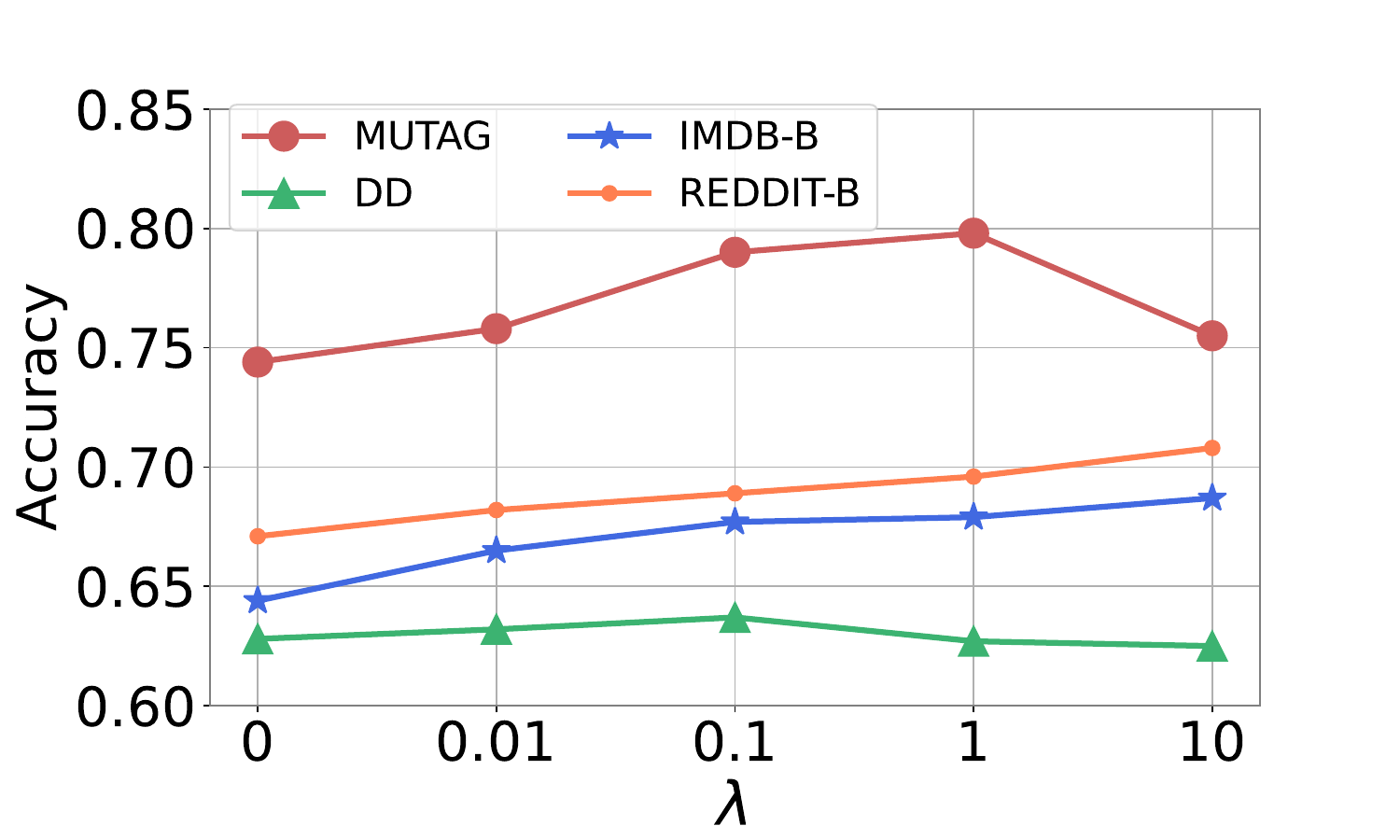}
    \end{minipage}
}
    \subfigure[Effects of $\lambda$]{
    \begin{minipage}[t]{0.23\linewidth} 
    \includegraphics[width=3.7cm]{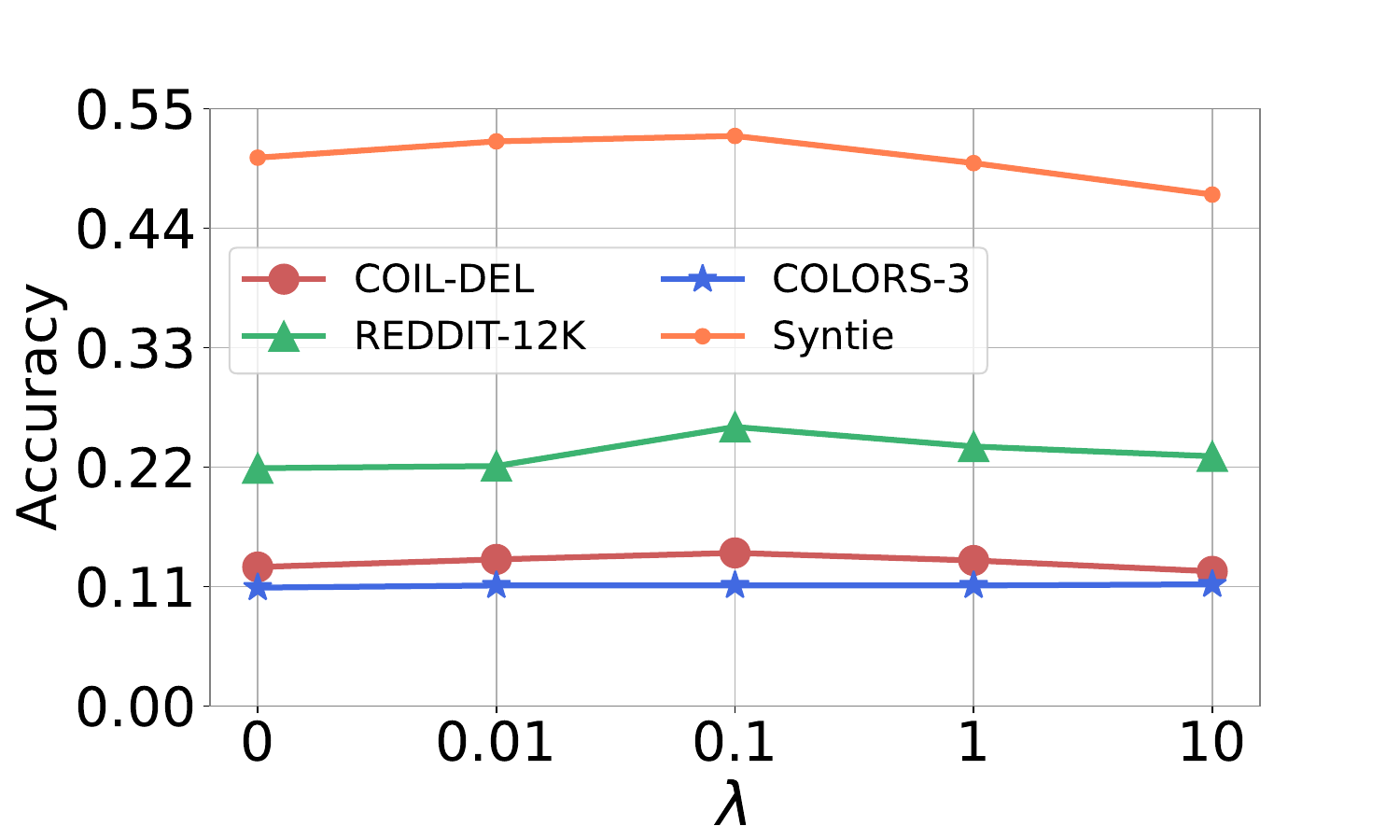}
    \end{minipage}
}
      \vspace{-0.2cm}
	  \caption{The hyper-parameters of $\epsilon$ and $\lambda$ on datasets.}
	  \vspace{-0.2cm}
	  \label{fig:paras} 
\end{figure*}

\paragraph{Effect of hyper-parameter (RQ3).}
We now study the effects of hyper-parameters $\epsilon$ and $\lambda$ on model performance with ACC and NMI (reported in Appendix D.4).
We first study the effects of $\epsilon$ by varying it in $\{0, 0.05,0.1,0.2,0.5\}$. The results are reported in Fig.\ref{fig:paras}(a)-(b).
Fig.\ref{fig:paras}(a)-(b) show that the performance improves when $\epsilon$ increases, then decreases or keeps a relatively stable level after $\epsilon$ reaches $0.1$.
We can conclude that when $\epsilon$ is too small, the augmentation consensus is too weak and the obtained pseudo labels cannot filter noise effectively.
When $\epsilon$ is too large, the augmentation consensus is too strong and the obtained pseudo labels may lose some useful information, which also reduces the clustering performance.
Empirically, we choose $\epsilon=0.1$.
Then we perform experiments by varying $\lambda$ in $\{0, 0.01,0.1,1,10\}$, the results are shown in Fig.\ref{fig:paras}(c)-(d).
From them, we can see that performance climbs up and then declines as $\lambda$ increases.
We can conclude that when $\lambda$ is too small, the instance uniformity is not enough and the overlapped clusters cannot be separated, limiting the clustering ability of \modelname.
When $\lambda$ becomes too large, the instance uniformity will suppress the clustering ability of \modelname, i.e., similar instances cannot be aggregated, leading to bad intra-cluster compactness.
With proper $\lambda$, the clustering ability of \modelname~can be fully exploited, thus enhancing the clustering performance.
Empirically, we choose $\lambda=0.1$ for DD, COIL-DEL, REDDIT-12K, Syntie, $\lambda=1$ for MUTAG, and $\lambda=10$ for IMDB-B, REDDIT-B, COLORS-3.

\section{Related work}

\subsection{Graph representation learning}
Graph-level representation learning has been widely studied in recent years in an unsupervised manner.
Existing work on this is mainly of two types, i.e., \textit{graph kernel methods} and \textit{contrastive methods}.
Constructing graph kernels is a traditional task in learning graph representations.
%
%
%
The researches are devoted to deciding suitable sub-structures and defining hand-crafted similarity measures between substructures \cite{shervashidze2009efficient,borgwardt2005shortest,shervashidze2011weisfeiler}.
On large datasets, the hand-crafted features of graph kernel methods lead to high dimensional, sparse or non-smooth representations and thus result in poor generalization performance \cite{narayanan2017graph2vec,suninfograph}.
Contrastive methods can be categorized into two groups.
One contrasts local and global representations to encode useful information \cite{velickovic2019deep,suninfograph}.
Another constructs graph augmentations to contrast two views of graphs \cite{you2020graph,chu2021cuco,you2021graph,li2022let,xia2022simgrace}.
%
%
%
%
%
%
%
Based on the observation of SimGRACE \cite{xia2022simgrace} that graph data can preserve their semantics well during encoder perturbations, we adopt perturbed encoder to obtain augmented views.
As a contrast, we adopt the same perturbation distribution all the time for simplicity's sake, while SimGRACE adopt different perturbation distribution for different layers of the GNN encoder.

\subsection{Graph clustering}
%
%
Graph clustering can be organized into two categories: node-level clustering and graph-level clustering.
Node-level clustering \cite{bo2020,tu2021deep,liu2022deep,liu2022hard}  aims to group nodes in a single graph into clusters.
SDCN \cite{bo2020} and DFCN \cite{tu2021deep} are proposed to jointly learn an auto-encoder and a graph auto-encoder in a united framework to alleviate the over-smoothing problem.
To learn more discriminative representations, DCRN \cite{liu2022deep} introduces a dual information correlation reduction mechanism.
%
Graph-level clustering aims to group multiple graphs into clusters.
Unlike the popularity of node-level clustering, graph-level clustering remains largely unexplored.
It is intuitive to utilize two-stage methods, i.e., graph-level representation learning methods \cite{shervashidze2009efficient,shervashidze2011weisfeiler,you2020graph,li2022let} followed by k-means for graph-level clustering.
But they cannot learn clustering-friendly representations and thus limiting the clustering performance.
There exists only one deep graph-level clustering method (i.e., GLCC \cite{ju2022glcc}), which utilizes cluster-level contrastive learning and pseudo labeling for obtaining cluster-friendly representations, achieving better performance.
However, it tends to obtain degenerate solutions that many samples are assigned to few clusters and consequently compromising the clustering performance \cite{zhou2022comprehensive}.
As a contrast, our method provide cluster-level uniformity for features on unit hypersphere to deal with the clustering degeneracy problem.

\section{Conclusion}
In this paper, we propose Uniform Deep Graph Clustering (\modelname), which includes \textit{pseudo label generation module} and \textit{representation enhancement module}, for graph-level clustering.
In the pseudo label generation module, we innovatively propose Augmentation-Consensus Optimal Transport (ACOT) to generate robust and uniformly distributed pseudo labels.
In the representation enhancement module, we introduce instance-based representation enhancement with contrastive learning and center-based representation enhancement with Center Alignment Optimal Transport (CAOT) to learn cluster-friendly representations while promoting cluster-level uniformity.
%
%
Extensive experiments conducted on eight popularly used datasets demonstrate the superior performance of our proposed \modelname.
\appendix
\section{ACOT}
As mentioned in Section 2.3, the ACOT problem is formulated as:
\begin{equation}
    \begin{aligned}
        &\min_{\bm{\pi},\bm{\pi}^{'}}{\langle \bm{\pi}, \bm{M} \rangle}+\epsilon {\rm{KL}(\bm{\pi}||\bm{\pi}^{'})}+\langle \bm{\pi}^{'},\bm{M}^{'}\rangle +\epsilon{\rm{KL}(\bm{\pi}^{'}||\bm{\pi})}\\
        &s.t.\,\,\bm{\pi}\bm{1}=\frac{1}{N}\bm{1},\bm{\pi}^T\bm{1}=\frac{1}{C}\bm{1},\bm{\pi}\geq 0,
        \bm{\pi}^{'}\bm{1}=\frac{1}{N}\bm{1},\bm{\pi}^{'T}\bm{1}=\frac{1}{C}\bm{1},\bm{\pi}^{'}\geq 0,
    \end{aligned}
\end{equation}
where $\epsilon$ is a hyper-parameter.
We first fix $\bm{\pi}^{'}$ and adopt the Lagrangian multiplier algorithm to optimize the problem:
\begin{equation}
    \begin{aligned}
        \min_{\bm{\pi}}{\langle \bm{\pi}, \bm{M} \rangle}+\epsilon {\rm{KL}(\bm{\pi}||\bm{\pi}^{'})}
        +\bm{f}^T(\bm{\pi}\bm{1}-\avector)+\bm{g}^T(\bm{\pi}^T\bm{1}-\bvector),
    \end{aligned}
    \label{eq:cuotpi}
\end{equation}
where $\bm{f}$, $\bm{g}$ are Lagrangian multipliers.
Taking the differentiation of Eq.\ref{eq:cuotpi} on the variable $\bm{\pi}$, we can obtain:
\begin{equation}
    \begin{aligned}
        \pi_{ij}=\pi^{'}_{ij}\exp(\frac{-M_{ij}-f_i-g_j-\epsilon}{\epsilon})>0.
    \end{aligned}
    \label{eq:updatepi}
\end{equation}
Due to the fact that $\bm{\pi}\bm{1}=\avector$ and $\bm{\pi}^T\bm{1}=\bvector$, we can get:
\begin{equation}
    \begin{aligned}
        \exp(-\frac{f_i}{\epsilon})=\frac{1}{N\sum_j \pi^{'}_{ij}\exp(\frac{-M_{ij}-g_j-\epsilon}{\epsilon})},
    \end{aligned}
    \label{eq:updatef}
\end{equation}
\begin{equation}
    \begin{aligned}
        \exp(-\frac{g_j}{\epsilon})=\frac{1}{C\sum_i \pi^{'}_{ij}\exp(\frac{-M_{ij}-f_i-\epsilon}{\epsilon})}.
    \end{aligned}
    \label{eq:updateg}
\end{equation}
With updated $\bm{f}$ and $\bm{g}$, we can obtain updated $\bm{\pi}$ through Eq.\ref{eq:updatepi}.
Then we fix $\bm{\pi}$ and adopt the Lagrangian multiplier algorithm to optimize the problem:
\begin{equation}
    \begin{aligned}
        \min_{\bm{\pi}^{'}}\langle \bm{\pi}^{'},\bm{M}^{'}\rangle +\epsilon{\rm{KL}(\bm{\pi}^{'}||\bm{\pi})}
        +{\bm{f}^{'}}^T(\bm{\pi}^{'}\bm{1}-\avector)+{\bm{g}^{'}}^T(\bm{\pi}^{'T}\bm{1}-\bvector),
    \end{aligned}
\end{equation}
where $\bm{f}^{'}$, and $\bm{g}^{'}$ are Lagrangian multipliers. Like the optimization of Eq.\ref{eq:cuotpi}, we can obtain:
\begin{equation}
    \begin{aligned}
        \pi^{'}_{ij}=\pi_{ij}\exp(\frac{-M^{'}_{ij}-f^{'}_i-g^{'}_j-\epsilon}{\epsilon})>0,
    \end{aligned}
    \label{eq:updategamma}
\end{equation}
\begin{equation}
    \begin{aligned}
        \exp(-\frac{f^{'}_i}{\epsilon})=\frac{1}{N\sum_j \pi_{ij}\exp(\frac{-M^{'}_{ij}-g^{'}_j-\epsilon}{\epsilon})},
    \end{aligned}
    \label{eq:updatefp}
\end{equation}
\begin{equation}
    \begin{aligned}
        \exp(-\frac{g^{'}_j}{\epsilon})=\frac{1}{C\sum_i \pi_{ij}\exp(\frac{-M^{'}_{ij}-f^{'}_i-\epsilon}{\epsilon})}.
    \end{aligned}
    \label{eq:updategp}
\end{equation}

In short, through iteratively updating Eq.\ref{eq:updatef}, Eq.\ref{eq:updateg}, Eq.\ref{eq:updatepi}, Eq.\ref{eq:updatefp}, Eq.\ref{eq:updategp} and Eq.\ref{eq:updategamma}, we can obtain the transport matrix $\bm{\pi}$ on Eq.\ref{eq:updatepi}. 
We show the iteration optimization scheme of ACOT in Algorithm 1.

\begin{minipage}{6.5cm}
\begin{algorithm}[H]
\begin{algorithmic}[1]
{
\STATE{\textbf{Input:} The cost matrixes $\bm{M}$ and $\bm{M}^{'}$.}
\STATE{\textbf{Output:} The transport matrix: $\bm{\pi}$.}
\STATE{\textbf{Procedure}:}
\STATE{Initialize $\bm{f}$, $\bm{g}$, $\bm{f}^{'}$, and $\bm{g}^{'}$ randomly;}
\STATE{Initialize $\pi^{'}_{ij}=1/(N\times C)$.}
\FOR{$i=1$ to $T$}
\STATE{Update $\bm{f}$ by Eq. (\ref{eq:updatef});}
\STATE{Update $\bm{g}$ by Eq. (\ref{eq:updateg});}
\STATE{Update $\bm{\pi}$ by Eq. (\ref{eq:updatepi});}
\STATE{Update $\bm{f}^{'}$ by Eq. (\ref{eq:updatefp});}
\STATE{Update $\bm{g}^{'}$ by Eq. (\ref{eq:updategp});}
\STATE{Update $\bm{\pi}^{'}$ by Eq. (\ref{eq:updategamma}).}
\ENDFOR
\STATE{Calculate $\bm{\pi}$ by Eq. (\ref{eq:updatepi}).}
}
\end{algorithmic}
\caption{The scheme of ACOT.}
\end{algorithm}
\end{minipage}
\begin{minipage}{7.5cm}
\begin{algorithm}[H]
\begin{algorithmic}[1]
{
\STATE{\textbf{Input:} The representations $\bm{X}$ and agents $\bm{W}$.}
\STATE{\textbf{Output:} The loss $\mathcal{L}_{center}$.}
\STATE{\textbf{Procedure}:}
\STATE{Initialize $\bm{f}_1$, $\bm{g}_1$, $\bm{f}_2$, $\bm{g}_2$ and $\bm{\mu}$ randomly;}
\FOR{$i=1$ to $T_1$}
\STATE{Update $\bm{f}_1$ by Eq. (\ref{eq:updatef1}); Update $\bm{g}_1$ by Eq. (\ref{eq:updateg1});}
\STATE{Update $\bm{\xi}$ by Eq. (\ref{eq:updatexi}); Update $\bm{\mu}$ by Eq. (\ref{eq:update_mu});}
\ENDFOR
\STATE{Normalize $\bm{\mu}$ to obtain $\bm{r}$;}
\FOR{$i=1$ to $T_2$}
\STATE{Update $\bm{f}_2$ by Eq. (\ref{eq:updatef2});}
\STATE{Update $\bm{g}_2$ by Eq. (\ref{eq:updateg2});}
\ENDFOR
\STATE{Update $\bm{\psi}$ by Eq. (\ref{eq:updatepsi});}
\STATE{Calculate $\mathcal{L}_{center}$ by Eq. (\ref{eq:losscenter}).}
}
\end{algorithmic}
\caption{The scheme of CAOT.}
\end{algorithm}
\end{minipage}

\section{CAOT}
As mentioned in Section 2.4.2, CAOT consists of \textit{centers discovery} and \textit{center alignment}.
The center discovery problem is formulated as:
\begin{equation}
    \begin{aligned}
        &\min_{\bm{\xi},\bm{\mu}}{\sum_{i=1}^{N}\sum_{j=1}^{C}\xi_{ij}||\bm{x}_i-\bm{\mu}_j||_2^2+\eta\bm{\xi}\log\bm{\xi}}
        \,\,s.t.\,\,\bm{\xi}\bm{1}=\frac{1}{N}\bm{1},\bm{\xi}^T\bm{1}=\frac{1}{C}\bm{1},\bm{\xi}\geq 0,
    \end{aligned}
\end{equation}
where $\bm{\xi}$ is the transport matrix, $\eta$ is a hyper parameter.
We adopt the Lagrangian multiplier algorithm to optimize the problem:
\begin{equation}
    \begin{aligned}
        \min_{\bm{\xi},\bm{\mu}}{\sum_{i=1}^{N}\sum_{j=1}^{C}\xi_{ij}||\bm{x}_i-\bm{\mu}_j||_2^2+\eta\bm{\xi}\log\bm{\xi}}+\bm{f}^T_1(\bm{\xi}\bm{1}-\avector)+\bm{g}^T_1(\bm{\xi}^T\bm{1}-\bvector),
    \end{aligned}
    \label{eq:center_discovery}
\end{equation}
where $\bm{f}_1$ and $\bm{g}_1$ are Lagrangian multipliers.
We first fix $\bm{\mu}$, and take the differentiation of Eq. (\ref{eq:center_discovery}) on the variable $\bm{\xi}$, we can obtain:
\begin{equation}
    \begin{aligned}
        \xi_{ij}=\exp(\frac{-f_{1i}-g_{1j}-||\bm{x}_i-\bm{\mu}_j||_2^2-\eta}{\eta})>0.
    \end{aligned}
    \label{eq:updatexi}
\end{equation}
Due to the fact that $\bm{\xi}\bm{1}=\frac{1}{N}\bm{1}$ and $\bm{\xi}^T\bm{1}=\frac{1}{C}\bm{1}$, we can get:
\begin{equation}
    \begin{aligned}
        \exp(-\frac{f_{1i}}{\eta})=\frac{1}{N\sum_j \exp(\frac{-||\bm{x}_i-\bm{\mu}_j||_2^2-g_{1j}-\eta}{\eta})},
    \end{aligned}
    \label{eq:updatef1}
\end{equation}
\begin{equation}
    \begin{aligned}
        \exp(-\frac{g_{1j}}{\eta})=\frac{1}{C\sum_i \exp(\frac{-||\bm{x}_i-\bm{\mu}_j||_2^2-f_{1i}-\eta}{\eta})}.
    \end{aligned}
    \label{eq:updateg1}
\end{equation}
Through performing Eq. (\ref{eq:updatef1}), Eq. (\ref{eq:updateg1}), and Eq. (\ref{eq:updatexi}), we can obtain $\bm{\xi}$.
Then we fix $\bm{\xi}$ and take differentiation of Eq. (\ref{eq:center_discovery}) on the variable $\bm{\mu}$, we can obtain:
\begin{equation}
    \begin{aligned}
        \bm{\mu}_j=\frac{\sum_{i=1}^N\xi_{ij}\bm{x}_i}{\sum_{i=1}^N\xi_{ij}}.
    \end{aligned}
    \label{eq:update_mu}
\end{equation}
In short, through iteratively updating Eq. (\ref{eq:updatef1}), Eq. (\ref{eq:updateg1}), Eq. (\ref{eq:updatexi}), and Eq. (\ref{eq:update_mu}), we can obtain the centers $\bm{\mu}$.
We normalize $\bm{\mu}$ and obtain hyperspherical centers, i.e., $\bm{r}=\frac{\bm{\mu}}{||\bm{\mu}||_2}$.

The center matching problem is formulated as:
\begin{equation}
    \begin{aligned}
        &\min_{\bm{\psi}}{\sum_{i=1}^{C}\sum_{j=1}^{C}\psi_{ij}(\exp(-\bm{w}_i\bm{r}_j^T))+\eta_1\bm{\psi}\log\bm{\psi}}
        \,\,s.t.\,\,\bm{\psi}\bm{1}=\frac{1}{N}\bm{1},\bm{\psi}^T\bm{1}=\frac{1}{C}\bm{1},\bm{\psi}\geq 0,
    \end{aligned}
\end{equation}
where $\bm{\psi}$ is the transport matrix, $\eta_1$ is a hyper parameter.
We adopt the Lagrangian multiplier algorithm to optimize the problem:
\begin{equation}
    \begin{aligned}
        \min_{\bm{\psi}}{\sum_{i=1}^{C}\sum_{j=1}^{C}\psi_{ij}(\exp(-\bm{w}_i\bm{r}_j^T))+\eta_1\bm{\psi}\log\bm{\psi}}+\bm{f}_2^T(\bm{\psi}\bm{1}-\avector)+\bm{g}_2^T(\bm{\psi}^T\bm{1}-\bvector),
    \end{aligned}
    \label{eq:match}
\end{equation}
where $\bm{f}_2$ and $\bm{g}_2$ are Lagrangian multipliers.
Taking the differentiation of Eq. (\ref{eq:match}) on the variable $\bm{\psi}$, we can obtain:
\begin{equation}
    \begin{aligned}
        \psi_{ij}=\exp(\frac{-\exp(-\bm{w}_i\bm{r}_j^T)-f_{2i}-g_{2j}-\eta_1}{\eta_1})>0.
    \end{aligned}
    \label{eq:updatepsi}
\end{equation}
Due to the fact than $\bm{\psi}\bm{1}=\frac{1}{N}\bm{1}$ and $\bm{\psi}^T\bm{1}=\frac{1}{C}\bm{1}$, we can get:
\begin{equation}
    \begin{aligned}
        \exp(-\frac{f_{2i}}{\eta_1})=\frac{1}{N\sum_j \exp(\frac{-\exp(-\bm{w}_i\bm{r}_j^T)-g_{2j}-\eta_1}{\eta_1})},
    \end{aligned}
    \label{eq:updatef2}
\end{equation}
\begin{equation}
    \begin{aligned}
        \exp(-\frac{g_{2j}}{\eta_1})=\frac{1}{C\sum_i \exp(\frac{-\exp(-\bm{w}_i\bm{r}_j^T)-f_{2i}-\eta_1}{\eta_1})}.
    \end{aligned}
    \label{eq:updateg2}
\end{equation}
That is, through iteratively updating Eq. (\ref{eq:updatef2}) and Eq. (\ref{eq:updateg2}), we can obtain the matching solution $\bm{\psi}$ on Eq. (\ref{eq:updatepsi}).
In the end, the center-agent representation enhancement loss can be defined as:
\begin{equation}
    \begin{aligned}
        \mathcal{L}_{center}={\sum_{i=1}^{C}\sum_{j=1}^{C}\psi_{ij}(\exp(-\bm{w}_i\bm{r}_j^T))}.
    \end{aligned}
    \label{eq:losscenter}
\end{equation}
We show the iteration optimization scheme of CAOT in Algorithm 2.

\section{Training Process}
We show the scheme of \modelname~in Algorithm 3.
\begin{algorithm}[H]
\label{alg:model}
\begin{algorithmic}[1]
{
\STATE{\textbf{Input:} Dataset $\bm{G}$.}
\STATE{\textbf{Output:} Clustering results $\bm{y}$.}
\STATE{\textbf{Procedure}:}
\STATE{Initialize agents $\bm{W}$ randomly.}
\FOR{$i=1$ to $t$}
\STATE{Update $\bm{\pi}$ by Algorithm 1 across the whole training process in a logrithmic distribution \cite{asano2020self}.} 
\STATE{Update $\bm{q}$ by $q_i=\arg\max_j\pi_{ij}$.}
\STATE{Update $\mathcal{L}_{center}$ by Algorithm 2.}
\STATE{Minimize the objective $\lambda\mathcal{L}_{instance}+\mathcal{L}_{agent}+\mathcal{L}_{center}$.}
\ENDFOR
\STATE{Calculate $\bm{y}$ by $y_i=\arg\max_j P_{ij}$.}
}
\end{algorithmic}
\caption{The scheme of \modelname.}
\end{algorithm}

\section{Experiment}

\subsection{Dataset}
Dataset statistics are reported in Table \ref{ta:dataset}, where L/S denotes the ratio of the largest cluster size to the smallest one.

\begin{table}
\caption{The statistics of the datasets. }
\label{ta:dataset}
\centering
\begin{tabular}{lcccccc}
\toprule
  Dataset & Category & \#Cluster & \#Graph & \#Node & \#Edge & L/S \\
\midrule
  MUTAG & Small molecules & 2 & 188 & 17.93 & 19.79 & 1.98\\
  DD & Bioinformatics & 2 & 1178 & 284.32 & 715.66 & 1.42\\
  COIL-DEL & Computer vision & 100 & 3900 & 21.54 & 54.24 & 1.00 \\
  IMDB-B & Social networks & 2 & 1000 & 19.77 & 96.53 & 1.00 \\
  REDDIT-B & Social networks & 2 & 2000 & 429.63 & 497.75 & 1.00 \\
  REDDIT-12K & Social networks & 11 & 11929 & 391.41 & 456.89 & 5.05 \\
  COLORS-3 & Synthetic & 11 & 10500 & 61.31 & 91.03 & 1.27\\
  Syntie & Synthetic & 4 & 400 & 95.00 & 172.93 & 1.22\\
\bottomrule
\end{tabular}
\end{table}

\subsection{Implemented details}
We adopt Graph Isomorphism Network (GIN) \cite{2019gin} as the encoder.
The number of GIN layers is selected from $\{3,5\}$, the hidden dimension is selected from $\{16,32,64\}$, the batch size is set to $128$, the temperature $\tau$ is set to $0.2$, the perturbation parameter $\sigma$ is selected from $\{0.1,1,2\}$.
We report the results from GLCC \cite{ju2022glcc} if available.
If results are not reported, we implement them and conduct a hyper-parameter search according to the original papers.
For graph kernel methods, we utilize the released code of TUDataset \cite{morris2020tudataset} to implement graph-level clustering.
For all deep graph representation learning methods, we utilize their respective released codes to implement graph-level clustering.
For GLCC, we implement the code based on the released code of GraphCL \cite{you2020graph}.

\subsection{Experimental results}
The average results of ARI on eight datasets are shown in Table \ref{tab:results_app}.

\begin{table}
\caption{ARI on eight datasets}
\label{tab:results_app}
\centering
\setlength{\tabcolsep}{0.35mm}{
\begin{tabular}{lcccccccc}
\toprule
& {MUTAG}
& {DD}
& {COIL-DEL}
& {IMDB-B}
& {REDDIT-B}
& {REDDIT-12K}
& {COLORS-3}
& {Syntie}\\

\midrule
Graphlet  & 0.000& -0.001& 0.033& 0.026& 0.000& -0.004& >1Day&0.038 \\ 
SP        & \textbf{0.424}& 0.001& 0.041& 0.044& 0.022& 0.005& >1Day&0.186 \\ 
WL        & 0.146& 0.002 & 0.037& 0.003& 0.021& 0.044& >1Day& \underline{0.365}\\
\midrule
InfoGraph & 0.221& -0.006& 0.046& 0.005& 0.000& 0.003& \underline{0.003}& 0.333\\
GraphCL   & 0.321& -0.009& 0.039& 0.008& 0.001& 0.021& 0.002&  0.323 \\
CuCo      & 0.268& -0.010& 0.046& 0.000& 0.000& 0.002& \underline{0.003}& 0.327\\
JOAO      & 0.291& -0.004& 0.041& 0.008& 0.001& 0.001& \underline{0.003}&  0.312\\
RGCL      & 0.117& -0.009& 0.045& 0.007& 0.001& 0.001& \textbf{0.004}& 0.330\\
SimGRACE  & 0.214& -0.003& \underline{0.047}& 0.007& 0.001& 0.005& \underline{0.003}& 0.241\\
\midrule 
GLCC      & 0.233& \underline{0.023}& \underline{0.047}& \underline{0.106}& \underline{0.087}& \underline{0.058}& 0.002& 0.295\\
\midrule
\modelname    &\underline{0.352} & \textbf{0.071} & \textbf{0.052}& \textbf{0.144}& \textbf{0.172}& \textbf{0.078}& 0.002& \textbf{0.466} \\
\bottomrule
\end{tabular}
}
\end{table}
\subsection{Effect of hyper-parameter}
\begin{figure*} 
    \centering
    
    \subfigure[Effects of $\epsilon$]{
    \begin{minipage}[t]{0.23\linewidth} 
    \includegraphics[width=3.7cm]{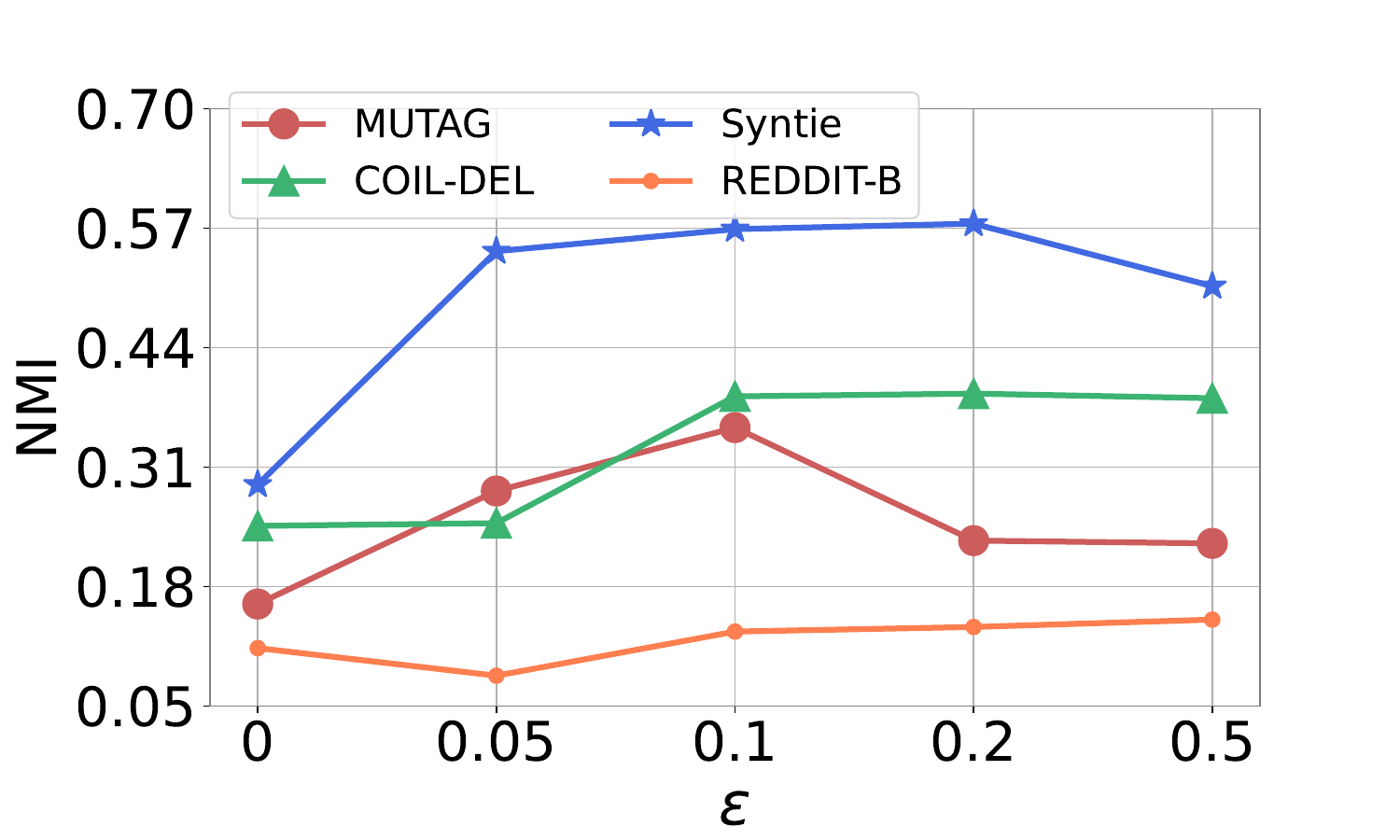}
    \end{minipage}
}
    \subfigure[Effects of $\epsilon$]{
    \begin{minipage}[t]{0.23\linewidth} 
    \includegraphics[width=3.7cm]{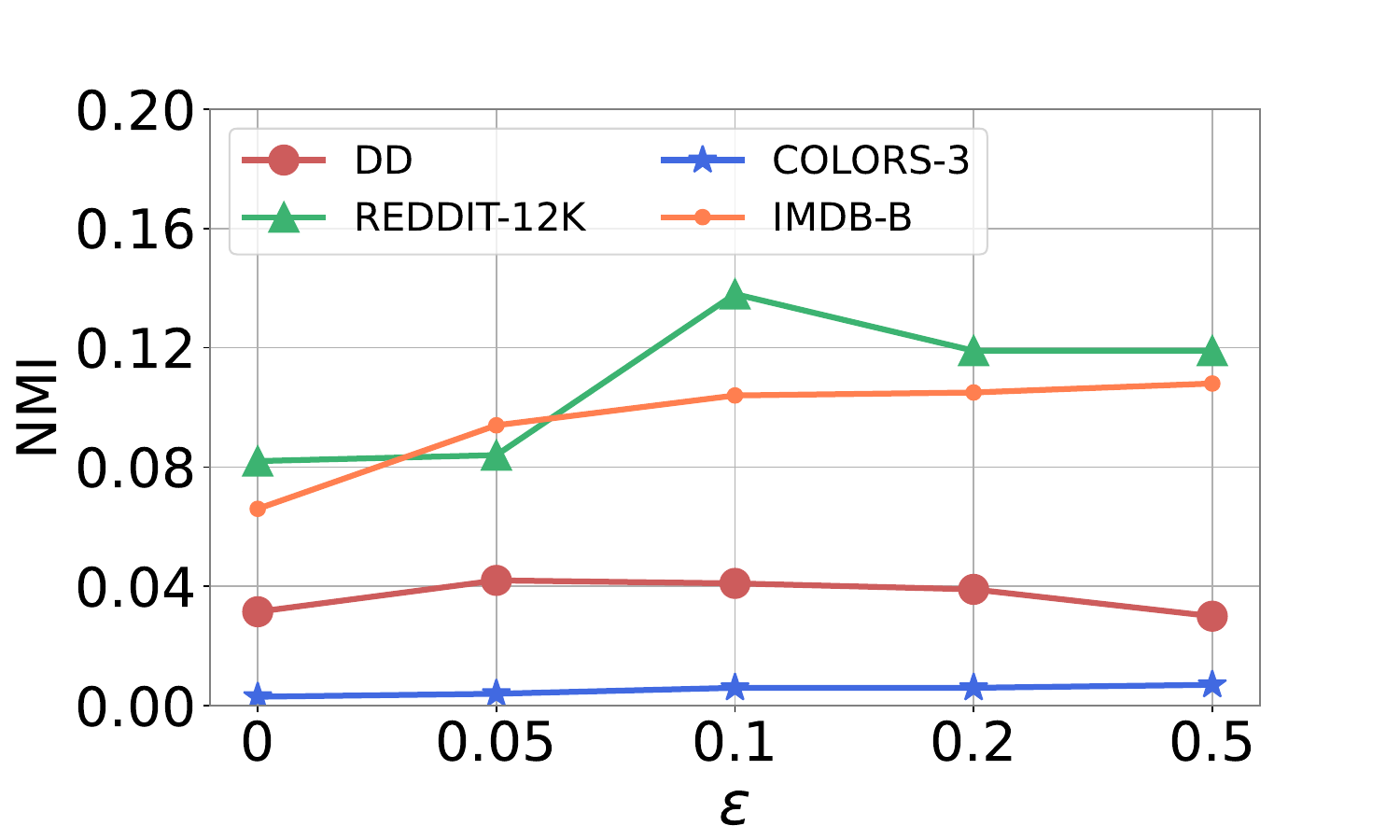}
    \end{minipage}
}
    \subfigure[Effects of $\lambda$]{
    \begin{minipage}[t]{0.23\linewidth} 
    \includegraphics[width=3.7cm]{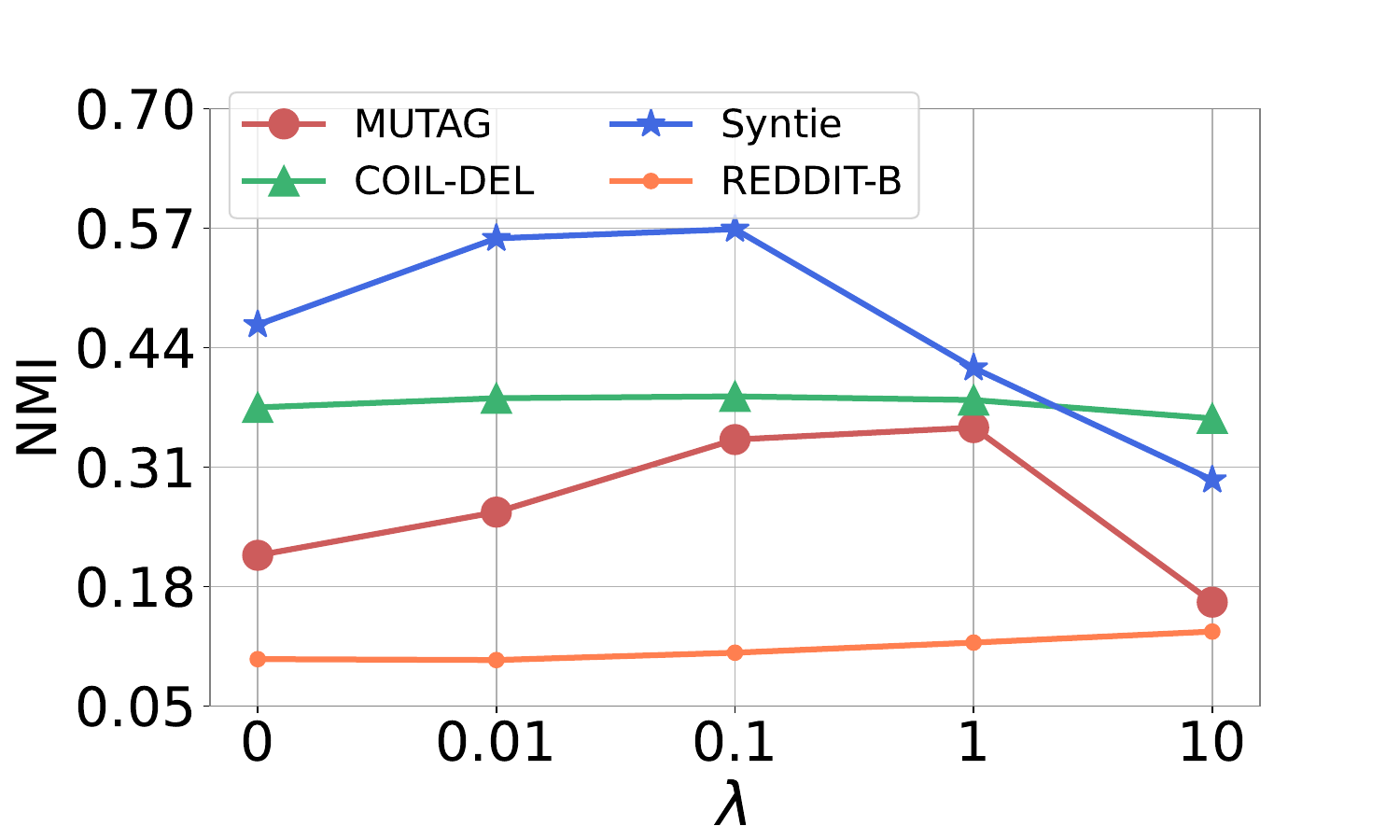}
    \end{minipage}
}
    \subfigure[Effects of $\lambda$]{
    \begin{minipage}[t]{0.23\linewidth} 
    \includegraphics[width=3.7cm]{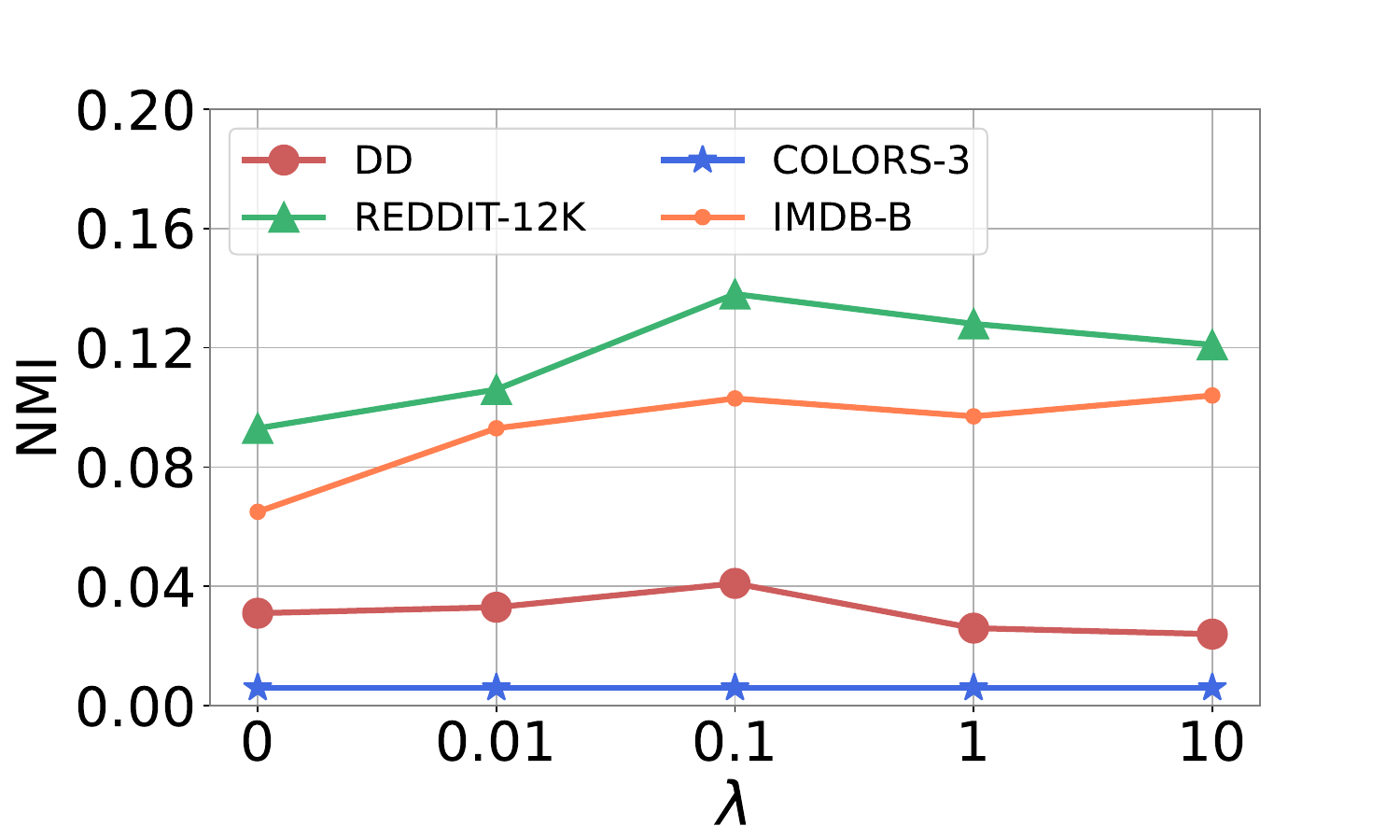}
    \end{minipage}
}
      \vspace{-0.2cm}
	  \caption{The hyper-parameters of $\epsilon$ and $\lambda$ on datasets.}
	  \vspace{-0.2cm}
	  \label{fig:paras_app} 
\end{figure*}

The effects of hyper-parameters $\epsilon$ and $\lambda$ on NMI are reported in Fig.\ref{fig:paras_app}(a)-(d).

\section{Limitations}
Like existing graph clustering methods, we assume the real cluster number is known.
In the future, we would like to explore a short text clustering method with an unknown number of clusters.
Moreover, the time complexity of augmentation-consensus optimal transport or center alignment optimal transport is $O(n^2)$, we are going to seek a new computation to reduce the complexity.


\bibliography{neurips}
\bibliographystyle{unsrt} 
\end{document}